\documentclass{article}

\usepackage[nonatbib,final]{neurips_2021}

\usepackage[utf8]{inputenc} 
\usepackage[T1]{fontenc}    
\usepackage{hyperref}       
\usepackage{url}            
\usepackage{booktabs}       
\usepackage{amsfonts}       
\usepackage{amsmath}
\usepackage{nicefrac}       
\usepackage{microtype}      
\usepackage{xcolor}         
\usepackage{mathptmx}   
\usepackage{pifont}
\usepackage{wrapfig}

\ifpdf
  \pdfoutput=1\relax                   
  \usepackage{graphicx}                
  \DeclareGraphicsExtensions{.pdf,.png,.jpg,.jpeg} 
\else
  \usepackage{graphicx}                
  \DeclareGraphicsExtensions{.eps}     
\fi%

\newcommand{\tool}{\textsc{Sim2RealViz}}
\newcommand{\gap}{sim2real gap}

\newcommand{\simu}{\color{black}\colorbox[rgb]{0.597,0.656,0.957}{\textsc{sim}}\color{black}}
\newcommand{\real}{\color{black}\colorbox[rgb]{0.957,0.656,0.597}{\textsc{real}}\color{black}}

\newcommand{\gt}{\color{black}\colorbox[rgb]{0.89,0.7,0.3}{\textsc{gt}}\color{black}}

\newcommand\myparagraph[1]{\textbf{#1} ---}
\usepackage{xspace}
\newcommand{\ie}{i.\,e.\xspace}

\newcommand{\eg}{e.\,g.,\xspace}

\usepackage{etoolbox}
\usepackage{enumitem}
\usepackage{amsmath}

\newcommand{\giturl}{{\small \url{https://github.com/Theo-Jaunet/sim2realViz}} }

\title{\tool: Visualizing the Sim2Real Gap\\ in Robot Ego-Pose Estimation}

\author{%
 \vspace{-.9cm}\\
  \textbf{Théo Jaunet}\\
  Liris, INSA-Lyon \\
   \And
    \vspace{-.9cm}\\
   \textbf{Guillaume Bono} \\
     Liris, INSA-Lyon \\
   \AND
    \vspace{-.5cm}\\
   \textbf{Romain Vuillemot} \\
     Liris, École Centrale de Lyon \\
   \And
    \vspace{-.5cm}\\
   \textbf{Christian Wolf} \\
     Liris, INSA-Lyon \\
}
\begin{document}

\maketitle


\begin{figure}[h!]
  \centering
  \vspace{-.9cm}
  \includegraphics[width=\linewidth]{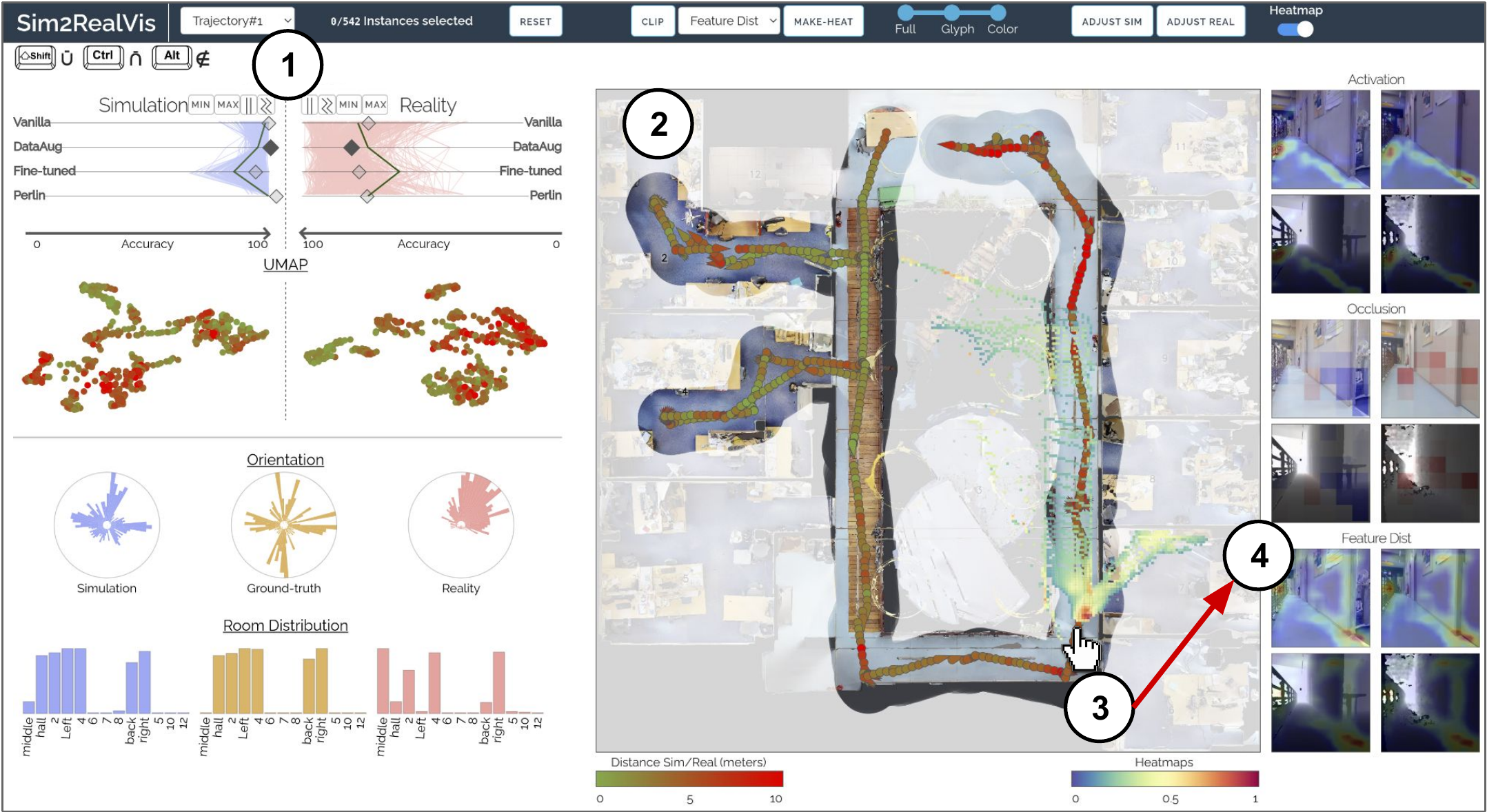}
  \caption{Using \tool, the \gap\ of a Data Augmentation model can be compared against other models (\eg Vanilla or Fine-tuned) and displayed on the real-world environment map  along with its performance metrics. In particular, \tool\ shows \ding{172} this model is particularly effective in simulation but we identified errors in the environment, such as the model failing to regress its position because of a closed-door that was opened in training. Such an error can then be selected by instance on the map \ding{173} to identify key features extracted by the model either as superimposed on the heat-map~\ding{174} or as a first-person view~\ding{175}.}
  \vspace{-.2cm}
  \label{fig:teaser}
\end{figure}

\begin{abstract}
The Robotics community has started to heavily rely on increasingly realistic 3D simulators for large-scale training of robots on massive amounts of data. But once robots are deployed in the real-world, the simulation gap, as well as changes in the real-world (e.g. lights, objects displacements) lead to errors. In this paper, we introduce \tool, a visual analytics tool to assist experts in understanding and reducing this gap for robot ego-pose estimation tasks, \ie the estimation of a robot's position using trained models. \tool\ displays details of a given model and the performance of its instances in both simulation and real-world. Experts can identify environment differences that impact model predictions at a given location and explore through direct interactions with the model hypothesis to fix it. We detail the design of the tool, and case studies related to the exploit of the regression to the mean bias and how it can be addressed, and how models are perturbed by vanishing landmarks such as bikes.

\end{abstract}

\section{Introduction}
\label{sec:intro}
Visual navigation is at the core of most autonomous robotic applications such as self-driving cars or service robotics. One of the main challenges for the robot is to efficiently explore the environment, to robustly identify navigational space, and eventually be able to find the shortest paths in complex environments with obstacles. The Robotics and Deep Learning communities have introduced models trained with Reinforcement Learning (RL), Inverse RL, or Imitation Learning, targeting complex scenarios requiring visual reasoning beyond waypoint navigation and novel ways to interact with robots, \eg combining vision, robotics, and natural language processing through queries like ``\emph{Where are my keys?}''. Current learning algorithms are not sampled efficiently enough, this kind of capability requires an extremely large amount of data. In the case of RL, this is in the hundreds of millions or in the billions of interactions --- this simply cannot be addressed in a reasonable amount of time using a physical robot in a real environment, which also may damage itself in the process.



To tackle this issue, the field heavily relies on simulation, where training can proceed significantly faster than in physical (wall clock) time on fast modern hardware, easily distributing multiple simulated environments over a large number of cores and machines. However, neural networks trained in simulated environments often perform poorly when deployed on real-world robots and environments, mainly due to the\emph{``Sim2Real gap''}, --- \ie the lack of accuracy and fidelity in simulating real-world environment conditions such as, among others, image acquisition conditions, sensors noise, but also furniture changes and other moved objects. The exact nature of the gap is often difficult to pinpoint. It is well known that adversarial examples, where only a few pixel shifts occur, considered as small artifacts by humans, or which might even be undetectable by humans, can directly alter the decisions of trained models~\cite{goodfellow2014explaining,moosavi2016deepfool,liu2018analyzing}. 

The \gap\ is currently addressed by various methods, including domain randomization, where the physical reality is considered to be a single parametrization of a large variety of simulations~\cite{tobin2017domain,mehta2020active}, and Domain Adaptation, \ie explicitly adapting a model trained in simulation to the real-world~\cite{tzeng2017adversarial,chadha2019improved}. However, identifying the sources of the sim2real gap would help experts in designing and optimizing transfer methods by directly targeting simulators and design choices of the agents themselves. To this end, we propose \tool, a visual analytics interface aiming to understand the gap between a simulator and a real-world environment. We claim that this tool is helpful to gather insights on the studied agents' behavior by comparing decisions made in simulation and in the real-world physical environment. \tool\ exposes different trajectories, and their divergences, in which a user can dive deeply for further analysis. In addition to behavior analysis, it provides features designed to explore and study the models' inner representations, and thus grasp differences between the simulated environment and the real-world as perceived by agents. Experts can rely on multiple-coordinated views, which can be used to compare model performances estimated with different metrics such as a distance, orientation, or a UMAP~\cite{mcinnes2018umap} projection of latent representations. In addition, experts dispose of three different approaches to highlight the estimated sim2real gap overlaid over either 3D projective inputs or over a bird's eye view (``Geo-map'') of the environment.
 
\tool\ targets domain experts, referred to as \emph{model builders and trainers}~\cite{Strobelt2017Lstmvis:Networks,Hohman2019VisualFrontiers}. The goal is assistance during real-world deployment, pinpointing root causes of decisions. Once a model is trained in simulation, those experts are often required to adapt it to real-world conditions through transfer learning and similar procedures. \tool\ provides information on the robot's behavior and has been designed to help end-users, building trust~\cite{dragan2013legibility, lipton2018mythos}. In Section~\ref{sec:case}, we report on insights gained through experiments using \tool, on how a selection of pre-trained neural models exploits specific sensor data, hints on their internal reasoning, and sensibility to sim2real gaps.

\section{Context and problem definition}

\begin{figure}[t!]
\centering
\includegraphics[width=\linewidth]{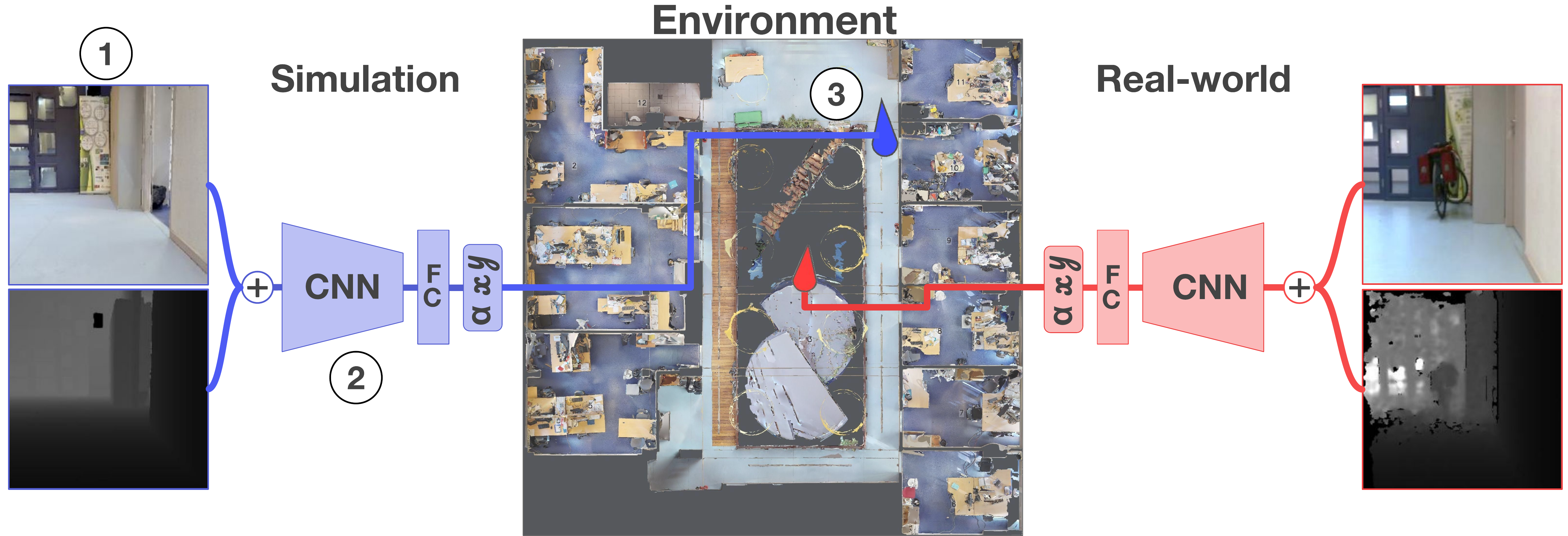}
\caption{In the studied robot ego-localization task, an RGB-D image~\ding{172}, is given to a trained model~\ding{173}, which uses it to regress the location ($x, y$), and orientation angle ($\alpha$) in the environment from which this image was taken from~\ding{174}. As illustrated above, images taken from the same coordinates in simulation and real-world~\ding{172} may lead to different predictions due to differences, such as here, among others, the additional presence of a bike in the scene. We are interested in reducing the gap between the \simu\ and \real\ predictions.}
\label{fig:gap}



\end{figure}

We study trained models for Ego-Localization of mobile robots in navigation scenarios, which regress the coordinates $(x,y)$ and camera angle $\alpha$ from observed RGB and depth images. Physical robots take these inputs from a depth-cam, whereas in the simulation they are rendered using computer graphics software from a 3D scan of the environment.
Fig.~\ref{fig:gap} provides a concrete example, where two images are taken at the same spatial coordinates~\ding{172}, one from simulation and the other from a physical robot. As our goal is to estimate the sim2real gap, we do not focus on generalization to unseen environments. Instead, our simulated environment corresponds to a 3D scanned version of the same physical environment in which the robot navigates, which allows precise estimation of the difference in localization performance, as gap leads to differences in predicted positions. The full extent of the gap, and how it may affect models is hard to understand by humans, which makes it difficult to take design choices and optimize decisions for sim2real transfer.




\myparagraph{Simulation} we use the Habitat~\cite{habitat19iccv} simulator and load a high fidelity 3D scan of a modern office building created with the Matterport3D software~\cite{Chang2017Matterport3D:Environments} from individual 360-degree camera images taken at multiple viewpoints. 
The environment is of size $22\times22$ meters and can potentially contain differences to the physical place due to estimation errors of geometry, texture, lighting, alignment of the individual views, as well as changes done after the acquisition, such as moved furniture, or opened/closed doors.
The Habitat simulator handles environment rendering and agent physics, starting with its shape and size (\eg a cylindrical with diameter 0.2m and height 1.5m), its action space (\eg turn left, right or move forward), and sensors --- a simulated \eg RGB-D camera. Depending on the hardware, the simulator can produce up to 10.000 frames per second, allowing to train agents on billions of interactions in a matter of days.

\myparagraph{Real-world} 
Our physical robot is a ``\emph{Locobot}''~\cite{locobot} featuring an RGB-D camera and an additional \textsc{lidar} sensor which we installed. We use the \textsc{lidar} and the \emph{ROS NavStack} to collect ground truth information on the robot's position $(x^*,y^*)$ and angle $\alpha^*$, used as a reference to evaluate ego-pose localization performances on the real-world. To increase precision, we do not build the map online with SLAM, but instead, export a global map from the 3D scan described above and align this map with the \textsc{lidar} scan using the \emph{ROS NavStack}.




\myparagraph{Ego-pose estimation: the trained agent}
Traditionally, ego-pose estimation of robots is performed from various inputs such as \textsc{lidar}, odometry, or visual input. Localization from RGB data is classically performed from keypoint-based approaches and matching/alignment \cite{brachmann2021limits}. More recently, this task has been addressed using end-to-end training of deep networks. 
We opted for the latter, and, inspired by poseNet~\cite{kendall2015posenet}, trained a deep convolutional network to take a stacked RGB-D image $X_i$ of shape $(256{\times}256{\times}4)$ and directly output a vector $Y_i=(x_i,y_i,\alpha_i)$ of three values: the coordinates $x_i,y_i$ and the orientation angle $\alpha_i$.
The model is trained on $60.000$ images sampled from the simulator with varying positions and orientations while assuring a minimal distance of $0.3$ meters between data points. We optimize the following loss function between predictions $Y_i=(x_i,y_i,\alpha_i)$ and ground truth (GT) $Y^*_i=(x^*_i,y^*_i,\alpha^*_i)$ over training samples $i$:
\begin{equation}
    \mathcal{L} = 
    (1-\gamma) 
    \sum_i 
    \left |
    \left | \ \
    \begin{bmatrix}  x \\ y \\ \end{bmatrix} -
    \begin{bmatrix}  x^* \\ y^* \\ \end{bmatrix}
    \ \
    \right |
    \right |_2^2   
    \ +
    \gamma 
    \sum_i 
    |\alpha -\alpha^*| \textrm{mod}~ 2\pi
\end{equation}
where $||.||_2$ is the $L_2$ norm and $\gamma$ is a weighting hyper-parameter set to 0.3 in our experiments.

\section{Related work}
\label{sec:related}

\vspace*{-3mm}
\myparagraph{Closing the sim2real gap and transfer learning}
Addressing the \gap\ relies on methods for knowledge transfer, which usually combine a large number of samples from simulation and/or interactions obtained with a simulator simulation with a significantly smaller number of samples collected from the real-world. Although machine learning is a primary way of addressing the transfer, it remains important to assess and analyze the main sources of discrepancies between simulation and real environments. A common strategy is to introduce noise to the agent state based on statistics collected in the real-world~\cite{habitat19iccv}. Additionally, tweaking the collision detection algorithm to prevent wall sliding has been shown to improve the performance in the real-world of navigation policies trained in simulation~\cite{kadian2020sim2real}, which tend to exploit inaccurate physics simulation. Another approach is to uniformly alter simulation parameters through domain randomization, \eg modifying lighting and object textures, to encourage models to learn invariant features during training~\cite{tobin2017domain,Tremblay_2018_CVPR_Workshops}. This line of work highly benefits from domain expert knowledge on the targeted environment, which can provide randomizations closer to reality~\cite{prakash2019structured,james2019sim}.

A different family of methods addresses the \gap\ through Domain Adaption, which focuses on modifying trained models' and their features learned from simulation to match those needed for high performance in real environments. This has been explored by different statistical methods from the machine learning toolbox, including discriminative adversarial losses~\cite{tzeng2017adversarial,chadha2019improved}. Feature-wise adaptation has also been addressed by extensive use of loss~\cite{fang2018multi,tzeng2020adapting}, and through fine-tuning~\cite{rusu2017sim}. Instead of creating invariant features, other approaches perform Domain Adaption at pixel level~\cite{bousmalis2017unsupervised,bousmalis2018using}. Despite great results, Domain Adaptation suffers from the need for real-world data, which is often hard to come by. We argue that there is a need for the assistance of domain experts and model builders to understand the main sources of sim2real gaps, which can then be leveraged for targeted adapted domain transfer, e.g. through specific types of representations or custom losses.

\myparagraph{Visual analytics and interpretability of deep networks} 
Due to their often generic computational structures, their extremely high number of trainable parameters (up to the orders billions) and the enormous amounts of data on which they have been trained, deep neural networks have a reputation of not being interpretable and providing predictions that cannot be understood by humans, hindering their deployment to critical applications. The main challenge is to return control over the decision process to humans, engineers, and model builders, which has been delegated to training data. This arguably requires the design of new tools capable of analyzing the decision process of trained models. The goal is to help domain experts to improve models, closing the loop, and build trust in end-users. These goals have been addressed by, both, the visualization, and machine learning communities~\cite{Hohman2019VisualFrontiers}.

Convolutional neural networks have been studied by exposing their gradients over input images, along with filters~\cite{zeiler2014visualizing}. Combined with visual analytics~\cite{liu2016towards}, this approach provided a glimpse on how sensible the neurons of those models are to different patterns in the input. Gradients can also be visualized combined with input images to highlight elements towards which the model is attracted to~\cite{Springenberg2014StrivingNetc}, with respect to the desired output (\eg a class).

More recently those approaches have been extended with class driven attributions of features~\cite{olah2018the, carter2019activation} which can also be formulated as a graph to determine the decision process of a model through interpretation of features (\eg black fur for bears)~\cite{hohman2020summit}. However, these approaches are often limited to image classification tasks such as ImageNet, as they need features to exploit human interpretable concepts from given images. 

\myparagraph{Interpretable robotics}
This classical line of work remains an under-explored challenge when applied to regression tasks such as robot ego-localization, our targeted application, in which attributions may be harder to interpret. To our knowledge, visualization of transfer learning, and especially targeting sim2real is an under-explored area, in particular for navigation tasks, where experiments with real physical robots are harder to perform compared to, say, grasping problems.
In~\cite{szabo2020visualizing}, the evolution of features is explored before and after transfer through pair-wise alignment. Systems such as~\cite{ma2020visual} address transfer gaps through multi-coordinated views and an in-depth analysis for models weights and features w.r.t. domains. In \cite{JaunetCGF2020}, the focus is on inspecting the latent memory of agents navigating in simulated environments.
Finally, common visualizations consist in heatmaps designed to illustrate results from papers in ML communities such as~\cite{tzeng2017adversarial, zhu2019sim}. 

Despite providing insights on how models adapt to different domains, and in contrast to our work, those methods have not been designed to directly target what parts of the environment, or which sensor settings may produce sim2real gaps, which we consider as relevant information for domain experts.

\section{\tool: A visual analytics tool to explore the sim2real gap}

We introduce \tool, an interactive visual analytics tool designed to assist domain experts in conducting in-depth analyses of the performance gaps between simulation and real environments of models whose primary task is ego-localization. The tool is implemented in JavaScript and the D3~\cite{Bostock2011D3:Documents} library to run in modern browsers and directly interacts with models implemented in Pytorch~\cite{NEURIPS2019_9015}. The tool and its source code are available as an open-source project at: \\
\giturl.

\subsection{Tasks analysis}
\label{sec:challenges}

Prior to the design of \tool, we conducted interviews with $3$ experts in Robotics and discussed their workflow, with the objective being to address and identify sim2real gaps. Two of those experts, co-author of this work, then took part in the design of \tool. The workflow of interrogated experts consisted in identifying failure cases through statistics or video replaying a robot's trajectory, and then manually finding equivalent images in simulation to compare to. From those discussions, and literature review introduced in sec.~\ref{sec:related}, we distill their process in the following three families of tasks.







\begin{enumerate}[label=\textbf{T\arabic*.},align=left]
\item \textbf{Fine-grained assessment of model performance gap between SIM and REAL} 
\label{C1} --- What is the best performing sim2real transfer method (\eg fine-tuning, domain randomization etc.)? What are the optimal hyper-parameters? Answering those questions requires experts to study a large number of predictions in SIM and REAL from a large number of observed images and evaluate performance distribution over different environment conditions and factors of variation.  




\item \textbf{Identification of the source of the performance gap}
\label{C2} --- what are the factors of variation in the environment, agent, or trained model, which are responsible for the performance gap? This is inherently difficult, as the sources may be due to the global environment (differences in e.g., lightening, 3D scanning performance), the agent (\eg differences in camera focal length or height) or changes due to the time span between scanning and physical deployment (\eg furniture changes). In addition,  some gaps may also be beyond human comprehension such as adversarial noise. For a human designer, it may not immediately be clear, which differences will have the largest impact on prediction performance.


\item\textbf{Closing the sim2real gaps}
\label{C3} --- successful knowledge transfer requires the simulator to be as close as possible to the real-world scenario with respect to the factors of variation identified in \ref{C2} The goal is to close the loop and increase prediction performance using the insights gained from using \tool.

\end{enumerate}


\subsection{Design rationale}
\label{sec:rationale}

Our design is centered around the comparison of simulation instances and real-world ones.
As we deal with complex objects, and because sim2real gaps can emerge from various sources, we implemented several views with different comparison strategies~\cite{gleicher_visual_2011}. As illustrated in Fig.~\ref{fig:teaser}, \tool\ follows the \emph{overview+detail} interface scheme~\cite{cockburn2009review} with a range from the most global views (left), to the most specific ones (right). To ease the comparison, simulation and real-world data are displayed next to each other within each view, with, if possible, simulation on the left side and real-world on the right side. The objective of the \emph{Statistics view}~(Fig.~\ref{fig:teaser}~\ding{172}) is to help in quickly identifying the performance of a model and to grasp global behavior with simple visualizations. The \emph{Geo-map}~(Fig.~\ref{fig:teaser}~\ding{173}), is key in providing context on the instance predictions, and for users to grasp what factors of variation may cause sim2real gaps. Finally, the \emph{Instance view}~(Fig.~\ref{fig:teaser}~\ding{174}), displays how models may perceive sim2real gaps under different scopes. To encode the main information related to the gap we used three colors corresponding to either \simu, \real, or \gt, accross the visualizations. We also used color to encode the distance between two sets of coordinates or the intensity of the models' attention towards parts of input images using a continuous turbo~\cite{turboscale} color scale, commonly used by experts, to emphasize the most critical instances, \ie those with high values.

\subsection{Main-stream workflow}

We now provide a typical workflow of use of \tool:

\begin{enumerate}

    \item Models are pre-loaded and their overall performances on both \simu\ and \real\ are displayed on the top left of \tool~(Fig.~\ref{fig:teaser}~\ding{172}).
    
    \item After model selection, users can start a fine-grained performance analysis of \textsc{sim} and \textsc{real} models by observing global statistics views such as a UMAP~\cite{mcinnes2018umap} projection of embeddings in which each dot is an instance, and its color encodes how far it is to its counterpart (sim or real). Followed by a radial bar chart of predicted or ground-truth orientation, and finally, a distribution of positions in which each room is a bar, and their height corresponds to how many predictions there is. In any of those views, users can select a set of instances to be inspected (\eg a cluster in UMAP).
    
    \item Any such selection updates a geo-map~(Fig.~\ref{fig:teaser}~\ding{173}), i.e. a ``geometric'' bird's eye view, in which users can inspect the predictions in a finer scale. Users can adapt the geo-map to either \emph{color}-mode which only displays ground-truth positions with their colors indicating how far sim and real predictions are, or \emph{full}-mode which displays sim predictions, ground-truth positions, and real predictions. Instances can be selected for further inspection by mouse-hovering them.
    
    \item An instance selection updates the instance view~(Fig.~\ref{fig:teaser}~\ding{174}) and displays heatmaps, which highlights the portions of images on which the model most focuses on, or which it perceives as different. Such a heatmap is also back-projected over the geo-map to highlight portions of the environment, which most likely carry sim2real gaps~(Fig.~\ref{fig:teaser}~\ding{175}).
\end{enumerate}

The views in \tool\ are multi-coordinated, i.e. any of them, including the geo-map, can be used as an entry point to formulate complex queries such as ``\emph{what are the instances which perform poorly in simulation, but good in real-world while being in a selected region of the environment?}''. Concretely, those combinations of selection can be done using sets operations (\emph{\{union, intersection, and complementary\}}), which can be selected through interactions with the corresponding views. This is further emphasized by the fact that the performance differences between simulation and real-world are also color-encoded on the geo-map. 

\subsection{Heatmaps}
\label{sec:heatmap}
To facilitate the inspection of the sim2real gap through image comparisons, \tool\ provides heatmaps superimposed over images, from a selected instance, to draw user attention towards key portions of inputs extracted by the trained model (Fig.~\ref{fig:teaser}~\ding{174}). Feature-wise visualizations are essential, as visual differences between simulated and real-world images perceived by humans may not correspond to differences in features with a high impact on model decisions. Fig.~\ref{fig:teaser}~\ding{174} illustrates the result of three approaches to generate those heatmaps, as follows (from top to bottom):

\myparagraph{Regression activation mapping} Inspired by grad-CAM~\cite{selvaraju2017grad} and RAM~\cite{wang2017diabetic}, we design heatmaps to highlight regions in the input, which have a high impact on model prediction. For each forward-pass of a model, we collect feature maps from the last CNN layer and multiply them by the weights of the last FC layer, obtaining an overlay of the size of the feature map, which is then re-scaled to fit the input image and normalized to fit a turbo color scale (Sec.~\ref{sec:rationale}). Similarity of activation maps between two similar \textsc{sim} and \textsc{real} images suggests a similarity of the two input images from the model's reasoning perspective.


   


\myparagraph{Sim/Real occlusion} Occlusion sensitivity~\cite{zeiler2014visualizing} is a common method to visualize how neural networks in computer vision rely on some portions of their input images. It consists in applying gray patches over an image, forwarding it to a model, and observing its impact on the model's prediction. By sampling a set of patches, we can then overlay the input with this information, blue color indicating that the occluded prediction is closer to the original ground truth, and red otherwise.

In our case, the intuition and solution are slightly different from the standard case. We are interested in areas of the input image, where the model performance is improved when the real-world observation is replaced by simulated information, indicating a strong sim2Real gap. We, therefore, occlude input REAL images with RGB or Depth patches from the corresponding simulated image. Thus, a further advantage of this approach is the possibility to discriminate between gaps in RGB or Depth input. The size of the patches is governed by a Heisenberg-like uncertainty trade-off between localization performance and measurement power. After experimenting with patch size ranging from 
$2\times2$ pixels to $128\times128$, we concluded that patches of $40\times40$ pixels, \ie a total of $6\times6$ patches per image, are the more suitable to analyze images on our computer as we estimated that response time for such an interaction should be less than one second. This is due to the fact that this is displayed on mouse-over, hence multiple instances can quickly be probed by an user, and a longer interaction time dampens the usability and user experience of \tool.

\myparagraph{Features map distance} Another approach implemented in \tool\ is to gather the feature map of the last CNN layer during a forward pass on both the simulation and its corresponding real-world image, and then compute a distance between them. The result is a matrix with the size of the feature map which is then overlaid like the activation mapping. After some iterations, we opted for the product of the cosine distance which favors small changes, and L1 which is more inclined to produce small spots. Such a product offers a trade-off between highlighting every change and face over-plotting while focusing only on one specific spot with the risk of losing relevant changes.


\begin{figure}[t!]
    \centering
    \includegraphics[width=\linewidth]{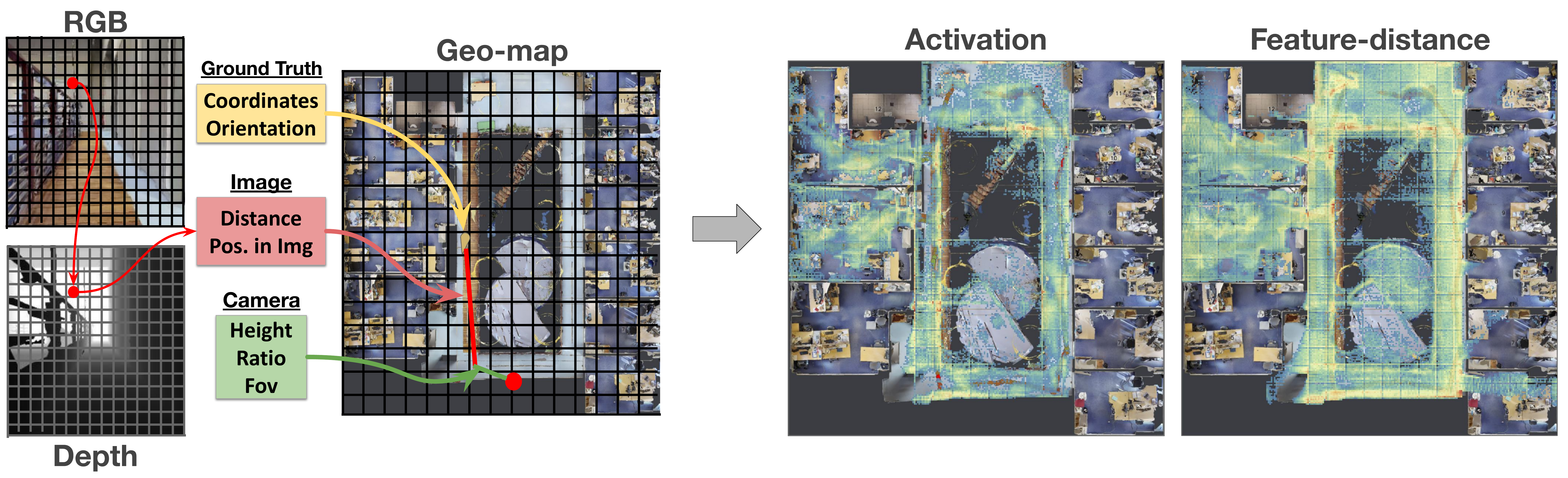}
    \caption{Conversion from pixels on a first-person point of view image to coordinates on a bird's eye geo-map (left). Such a process, used in \tool\ to display global heatmaps (right) on the geo-map, relies on ground-truth, image, and camera information. To optimize their computation, geo-maps are discretized into squares larger than a pixel, as a trade-off between the accuracy of projections, and the time to provide feedback to users upon interactions}
    \label{fig:global_heats}
\end{figure}

\subsection{Contextualization on the global geo-map}

As illustrated in Fig.~\ref{fig:teaser}~\ding{175} and in Fig.~\ref{fig:global_heats}, information from the individual first-person 3D projective input images, including heatmaps, can be projected into the global bird's eye view, and thus overlaid over the geo-map. This is possible thanks to ground truth information, \ie coordinates, and orientation of the instance, combined with information of the calibrated onboard cameras (simulated and real) themselves such as its field-of-view, position on the robot, resolution, and the range of the depth sensor. To do so, the environment is discretized in $264\times264$ blocks initially filled with zeroes, and images are downsampled to $128{\times}128$. Each cell is converted into ($x,y$) coordinates, and its average value from a heatmap is summed with the closest environment block to ($x,y$) coordinates. Finally, the values of environment blocks are normalized to fit the turbo color scale and then displayed as an overlay on the \emph{geo-map}. 
This process can also be applied to the complete dataset available at once to provide an overview of sim2real gaps of the environment as perceived by a model. Fig.~\ref{fig:global_heats} shows the conversion of heatmaps from the complete real-world dataset to a geo-map overlay using different aggregation strategies. This overlay can be displayed using the button \emph{make-heat} from the geo-map view~(fig~\ref{fig:teaser}~\ding{173}).



\subsection{Exploration of input configurations}
\label{sec:inputs}
\begin{wrapfigure}{r}{0.4\textwidth}
\centering
\includegraphics[width=5.5cm]{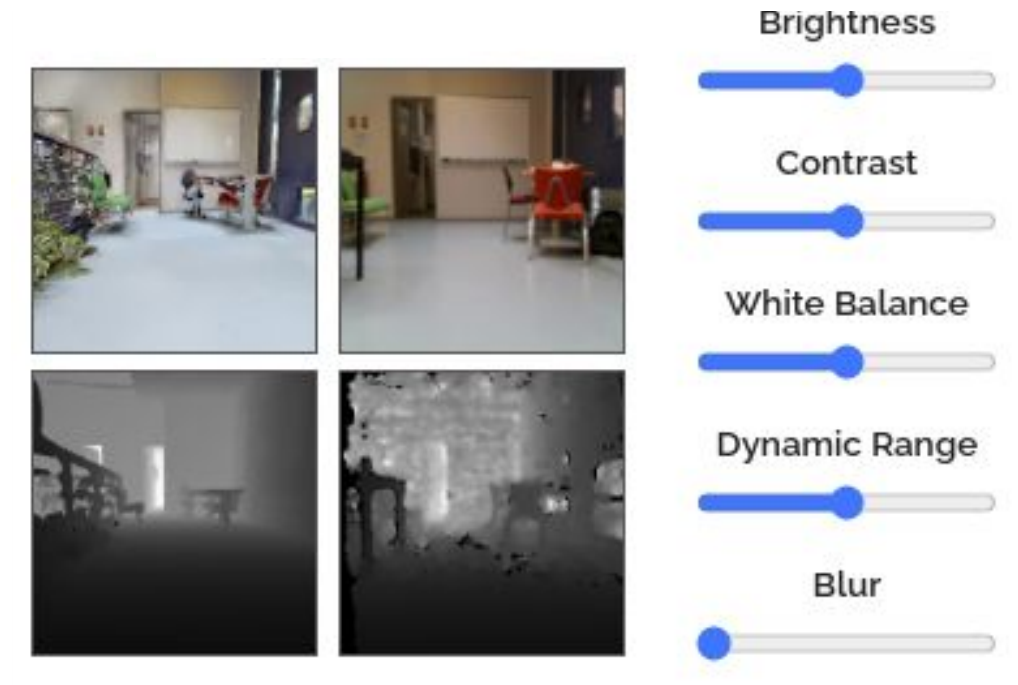}
\caption{By clicking on the \emph{adjust} button (on the top-right of Fig.~\ref{fig:teaser}), users can display sliders on the right of instance view Fig.~\ref{fig:teaser}~\ding{175}) that can be used to fine-tuning real-world images with filters and observe how it affect models' prediction.}
\label{fig:adjust}
\end{wrapfigure}

To check the impact of simulation gaps due to global imaging parameters, \tool\ provides ways to adjust real-world images through filters such as brightness, contrast, temperature, and dynamic range of depth. As illustrated in Fig.~\ref{fig:adjust}, those filters can are generated with sliders on the right of instance view (Fig.~\ref{fig:teaser}~\ding{175}). Any adjustment on a selected instance updates the corresponding prediction in real-time. Once a set of adjustments is validated by the user, it can be saved, applied to the whole real-world dataset, and treated as a new model in the~\emph{model gaps overview \ding{172}} for further analysis. 

The configuration of inputs can also be used on images from simulation, to analyze the performance of the model under specific Domain Randomization configurations, or simulation settings such as, for example, the height of the camera, or its FoV. Of course, to have an impact of the simulation on real-world images, the models need to be retrained outside of \tool. To do so, one must first generate a new dataset with the modified settings, in our case $60k$ images, which can take around half a hour as we also need to enforce diversity of sample images (coordinates and orientation). Then, train from scratch our model takes around $3$ to $4$ hours on our single NVIDIA Quadro P4000 GPU. Despite such a delay, adjusting simulation images in \tool\ can be useful to help manually extracting parameters of the real-world camera, and hence assist in the configuration of the simulator. Producing images, by configuring the simulator with direct feedback, should reduce the workload usually required to configure simulators and real-world robots.

\section{Case studies}
\label{sec:case}
We report on illustrative case studies we conducted to demonstrate how \tool\ can be used to provide insights on how different neural models may be influenced by sim2real gaps. During these experiments, \tool\ is loaded 
with the following methods for sim2real transfer: \emph{vanilla} (\ie no transfer, deployment as-is), \emph{dataAug} (\ie with Domain Randomization over brightness, contrast, dynamic range, hue), \emph{fine-tuning} on real-world images, and \emph{perlin}, a hand-crafted noise on depth images designed to be similar to real-world noise. We use visual data extracted from two different trajectories of the physical Locobot agent in the real environment performed with several months between them and at different times of the day, which provides a diverse range of sim2real gaps and optimizes generalization. Those models and data are available in our GitHub repository at: \giturl.

Insights on sim2real gaps grasped using \tool\ can be leveraged from two different perspectives echoing current sim2real transfer approaches. First, similar to Domain Adaptation, we can provide global modifications of the \real\ images (\eg brightness), which can be placed as filters and used in \tool. Second, related to Domain Randomization, by modifying the simulator settings (\eg adding or removing objects in the environment), and then by training a new model on it. In what follows, we describe several types of sim2real gaps, which have been identified and partially addressed in our experiments.

\begin{figure}[t!]
 \vspace{-.3cm}
    \centering
    \includegraphics[width=0.75\linewidth]{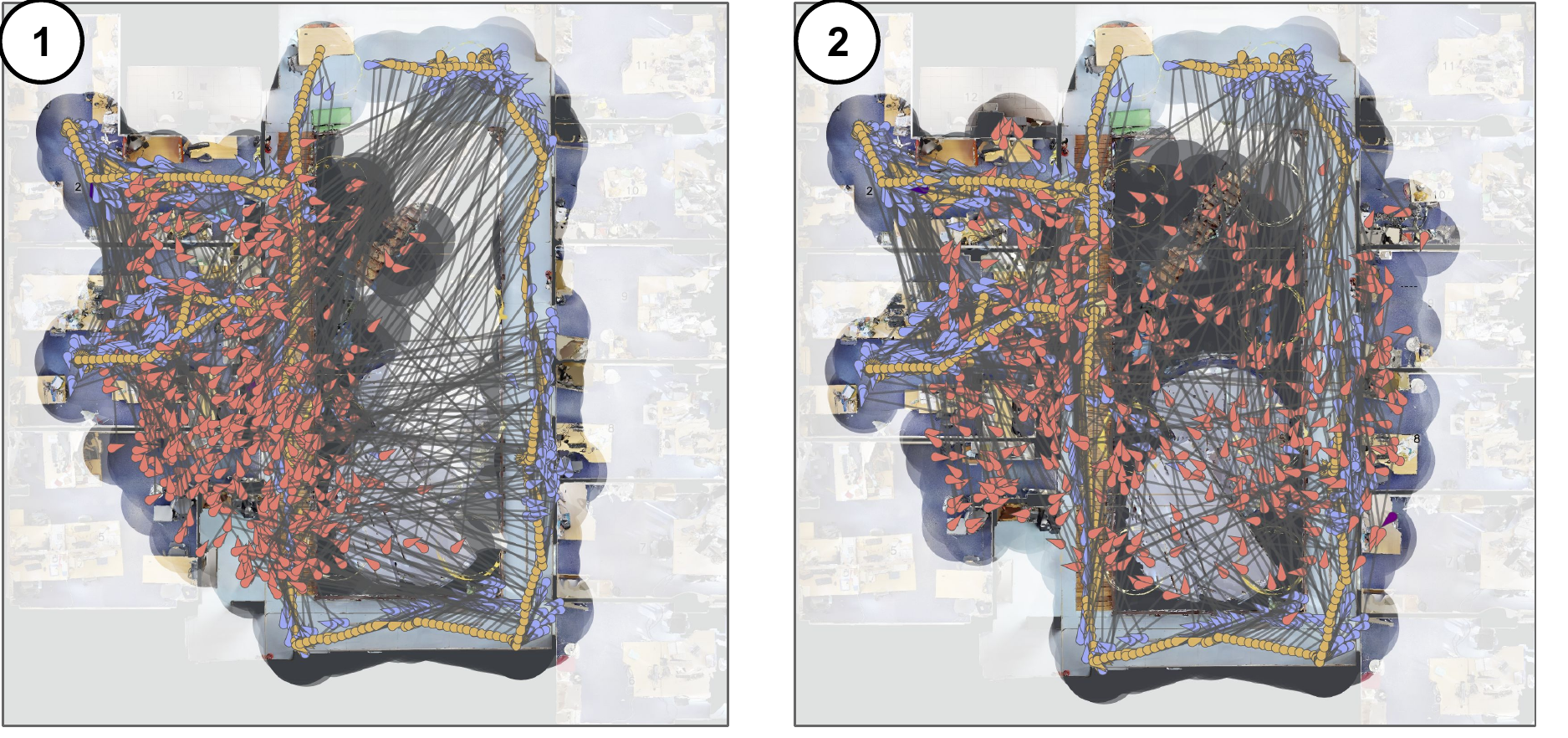}
    \caption{By using the \emph{full} encoding, we can observe that most real-world predictions are located in the half left of the environment~\ding{172}. Hence, instances sampled from the half right of the environment provide the worst predictions. However, when we slightly increase the brightness of each real-world image, we can observe that instances are more evenly distributed over the environment~\ding{173}.}
    \label{fig:case1}
 \vspace{-.4cm}
\end{figure}

\myparagraph{Unveiling biases in predictions}
Once loaded, users can observe how models perform on simulated and real-world data provided by different models trained and transferred with different methods, as shown in Fig.~\ref{fig:teaser}~\ding{172}. We report that best real-world performances are reached using \emph{dataAug}, with an average of 84\% accuracy, rather than \emph{Fine-tuning}, with an average accuracy of 80\%. This performance is evaluated on traj\#1, whereas traj\#2 had been used for fine-tuning on real-world data, ensuring generalization over experimental conditions in the environment. In what follows we will focus on the \emph{dataAug} model, which a user can further analyze by clicking on its corresponding diamond (Fig.~\ref{fig:teaser}~\ding{172}). This updates every other views to display data extracted from this model. To assess what the worst real-world prediction is, users can use the \emph{min} filter on the top of Fig.~\ref{fig:teaser}~\ding{172}. This removes from each view of \tool\ instances whose real-world performances are not among the bottom $15\%$. In our case, the remaining data displayed corresponds to instances sampled from the right corridor regardless of the model used. We conjecture, that corridors are among the most difficult instances as they are quite large and lack discriminative landmarks. However, in opposition, by using the \emph{max} filter, we can also observe that the left-side corridor is among the most successful predictions. By hovering those corridor instances with a successful transfer, we can inspect activation heatmaps and observe that model attention is driven towards the limit between a wooden path (unique to the left corridor) and a wall. Thus, the model seems to have learned to discriminate between corridors, which suggests that the confusion between them may be due to other reasons. By switching the encoding on the geo-map to \emph{full} using the slider on the middle top of \tool, the geo-map updates to display \simu, \real, and \gt positions (Fig.~\ref{fig:case1}~\ding{172}). With this, we can observe that the \emph{vanilla} model, incorrectly predicts real-world positions from the half right of the environment in the half left. Since those instances are correctly predicted in simulation, this indicates a very strong bias from most of the half right real-world instances. A similar phenomenon is also observed for the \emph{dataAug} model with instances on the right corridor creating predictions pointing to the middle of the environment, which is also an unreachable area.

\begin{wrapfigure}{r}{0.5\textwidth}
\centering
\vspace{-0.8cm}
\includegraphics[width=7cm]{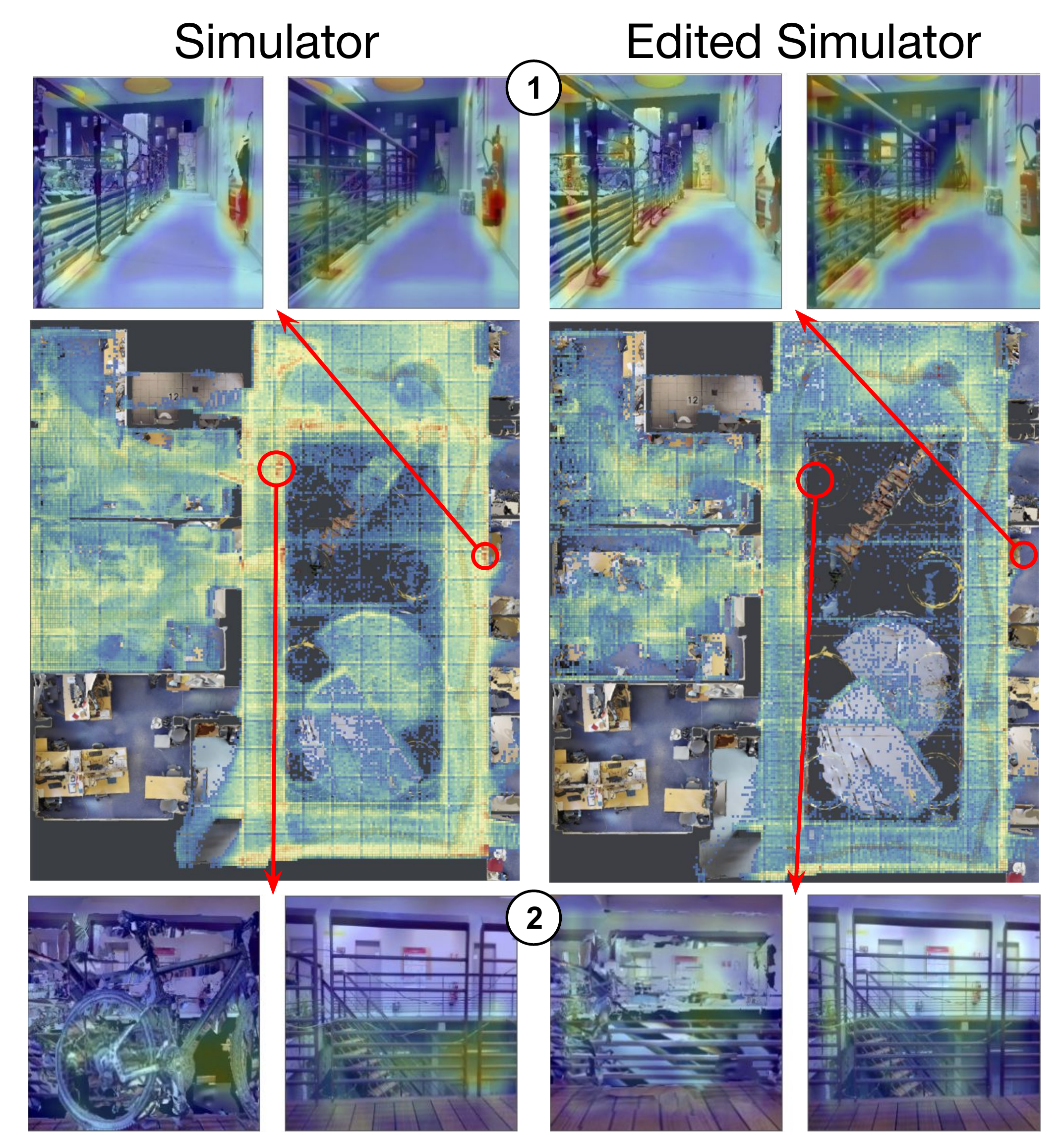}
\caption{With global heatmaps of \emph{feature-distance}, we can observe (in red) areas of the environment that may be affected by a sim2real gap. Those areas correspond to changes in objects present in the simulation, for instance as illustrated here, a fire-extinguisher. By removing such objects in simulation and retraining a model on them, we can observe that they disappeared from most highlighted areas.}
\label{fig:case2}
\vspace{-.5cm}
\end{wrapfigure}

\myparagraph{Closing the loop with filters}
We verify the hypothesis of regression to the mean, which is often an ``easy'' short-cut solution for a model in absence of regularities in data, or when regularities are not learned. The following focuses on the \emph{vanilla} model, as it is the one with the most real-world predictions on the half left of the environment. We perform global adjustments of the imaging parameters of the real-world images as described in Sec.~\ref{sec:inputs}, in particular setting both RGB and depth input to zero (\ie uniform black images), leading to the same constant predictions in the middle of the environment, corroborating the hypothesis.

While adjusting the brightness filter, we noticed that making images from the right corridor darker, yielded real-world predictions to be even more to the half left of the environment. In opposition, by making those images 15\% brighter, yielded real-world predictions, more accurately, in the half right of the environment leading to a slight improvement of the overall performance of 1.5\% (Fig~\ref{fig:case1}~\ding{173}).

\myparagraph{Sim2real gaps due to layout changes}
Trained models deployed to real environments need to be resilient to dynamic layout changes such as opened/closed doors, the presence of humans or moved furniture and other objects. In \tool, this can be investigated using the global heatmap with \emph{feat-dist}, displayable with the \emph{make heat} button on the middle top of \tool\, as seen in Fig.~\ref{fig:teaser}. In such geo-map overlay, some areas of the environment noticeably stand out (red in Fig.~\ref{fig:case2}). By hovering over instances on the geo-map nearby those areas, and browsing their corresponding images (as in Fig.~\ref{fig:teaser}~\ding{175}), we can observe that those areas are triggered by different factors. For instance, in Fig.~\ref{fig:case2}~\ding{173}, the highlighted area corresponds to the presence of a bike in the simulated data, which was not present when the real-world data had been captured. Other areas correspond to changed furniture, and imperfections of the simulation when rendering, for instance, a fire-extinguisher (Fig~\ref{fig:case2}~\ding{172}). Such behavior, which can be observed across models, may benefit from specific attention while addressing sim2real transfer.

\myparagraph{Editing the simulator}
In order to test such a hypothesis, we manually edited the simulator and removed objects corresponding to two red areas using \emph{blender}, a 3D modeling software. This new simulation is then used to generate $60k$ new images. Using these data, we trained a new \emph{dataAug} model and loaded its predictions in \tool\ to evaluate the influence of those changes on real-world performance. Fig.~\ref{fig:case2} shows the global \emph{feat-dist} heatmap on trajectory\#1 created with the new model, taking into account the changes. We can see that the areas with the most significant differences are more uniformly distributed over the environment. Since global heatmaps are normalized over the complete dataset, this indicates that, to the model, those areas are now closer in feature space.

The experiments described above have lead to a better understanding of the sim2real gap of our baseline agent, and we reported more robust localization performance once these insights were leveraged to modify the simulator or by learning filters for the existing model. We hope that \tool\ will be adopted and facilitate the design and training of trained robotic agents.

\vspace*{-4mm}
\section{Conclusion and perspectives}
\vspace*{-2mm}
We introduced \tool, an interactive visual analytics tool designed to perform an in-depth analysis of the emergence of sim2real gaps from neural networks applied to robot ego-localization. \tool\ supports both overview and comparison of the performances of different neural models, which instances can be browsed based on metrics such as performance or distribution. Those metrics can be combined using set operations to formulate more elaborated queries. We also reported scenarios of use of \tool\ to investigate how models are inclined to exploit biases, such as regression to the mean, and are easily disturbed by layout changes, such as moved objects.



\bibliography{sim2realViz}
\bibliographystyle{abbrv-doi}

\end{document}